\documentclass[12pt,fleqn]{article}
\usepackage[]{graphicx}
\begin{document}
\title{Filter Characteristics in Image Decomposition with Singular Spectrum Analysis}

\author{Kenji Kume \thanks{Department of Physics, Nara Women's University, Nara 630-8506, Japan. } \and Naoko Nose-Togawa \thanks{Research Center for Nuclear Physics, Osaka University, Ibaraki 567-0047, Japan}  }

\maketitle
\thispagestyle{empty}

\begin{abstract}
Singular spectrum analysis is developed as a nonparametric spectral decomposition of a time series.   It can be easily extended to the decomposition of multidimensional lattice-like data through the filtering interpretation. In this viewpoint, the singular spectrum analysis can be understood as the adaptive and optimal generation of the filters and their two-step point-symmetric operation to the original data. In this paper, we point out   that, when applied to the multidimensional data, the adaptively generated filters exhibit symmetry properties resulting from the bisymmetric nature of the lag-covariance matrices. The eigenvectors of the lag-covariance matrix are either symmetric or antisymmetric, and for the 2D image data, these lead to the differential-type filters with even- or odd-order derivatives.  The dominant filter is a smoothing filter,  reflecting the dominance of low-frequency components of the photo images. The others are the edge-enhancement or the noise filters corresponding to the 
band-pass or the high-pass filters. The implication of the decomposition to the image denoising is briefly discussed.
\end{abstract}

\newpage
\addtocounter{page}{-1}
\section{Introduction}
\label{intro}
Singular spectrum analysis (SSA) is a nonparametric and adaptive decomposition of a time series. In the SSA, the original time series is decomposed into arbitrary number of additive components and they are interpreted as the slowly varying trend, oscillatory and noise components, respectively [2,3]. In the conventional approach, the original time series $\{ x_i ; \ i=0,...,L-1 \}$ is embedded into the trajectory matrix $X$ through the lagged vectors of length $K$ by forming $L-K+1$ lagged vectors.
 Then the singular value decomposition of the trajectory matrix $X$ is carried out 
as 
\begin{equation}
X=X_1 + X_2+....+X_K \ ,
\end{equation}
with $X_j$ are the rank 1 matrices. After grouping the matrices $X_j$, the time series is reconstructed by projecting the matrices $X_j$ into Hankel form through the anti-diagonal averaging.
Recently, it has been shown that these SSA procedure can equivalently be interpreted as the two-step (forward and the reverse) filtering of a time series [6,8]. The eigenvector $ \{ v^{(k)} ; k=1,...,K \} $ of the matrix $X^TX$ can be interpreted as the convolution coefficients of the non-causal FIR filter. Let us denote the discrete Fourier transform of the eigenvector $\{ v^{(k)} \}$ for $X^TX$ as 
$\{ \hat{v}^{(k)}_\alpha \}$.  In Fourier space, the first (second)  step filtering can be expressed as the multiplication of $\hat{v}^{(k)*}_\alpha$ ($\hat{v}^{(k)}_\alpha$)  to the original sequences. Then the two-step filtering can be expressed as the multiplication of the zero-phase filters $|\hat{v}^{(k)}_\alpha|^2$ [7]. The normalization and the perfect-reconstruction 
properties can be expresses as 
\begin{equation}
{1 \over L} \sum_{\alpha=0}^{L-1} |\hat{v}^{(k)}_\alpha|^2 =1 \ ,
\end{equation}
and
\begin{equation}
{1 \over K} \sum_{k=1}^K |\hat{v}^{(k)}_\alpha|^2  =1 \ ,
\end{equation}
and these are the consequences of the normalization and the completeness properties of the eigenvectors $\{ v^{(k)} \}$ for $X^TX$ [8].

Though the filtering interpretation of the SSA algorithm is equivalent to the conventional treatment, it has several advantages. In the conventional approach, the singular value decomposition of the trajectory matrix and the reconstruction procedure through the Hankel projection seems to be quite different procedures and this makes it difficult to extend SSA to higher dimension.  With the filtering interpretation, these procedures can be seen as the forward and the reverse two-step point-symmetric convolution with respect to the reference points. This largely simplify the algorithms and enables us to  extend the SSA to the decomposition of multidimensional data with arbitrary dimension  [9].  Also, we can study the properties of the filters separately and it helps us for the detailed study of the spectral meaning of the decomposition. 

There have been several works in which the SSA algorithm has been applied to 2D image data [4,12] or three dimensional polygonal data [10,11].  In this paper, we focus on the SSA decomposition of the 2D image data.
Under the periodic boundary condition for the image data, the matrix $X^TX$ becomes to be the bisymmetric type and it leads to either the symmetric or the antisymmetric eigenvectors. A single step filtering generated with these eigenvectors 
can be interpreted as the differential operation to the images and the symmetric (antisymmetric) eigenvalues lead to the derivative filters of the even- ( odd- ) order.  
In the image processing, various differential filters are designed to detect and enhance the edge lines in the images.  In the SSA decomposition of the 2D image data, these differential filters are adaptively and optimally generated.  We have examined these points in detail, focusing our attention to the characteristics of the filters in the SSA.  Finally, we briefly discuss the possibility of the denoising with the SSA decomposition.

In Sec. 2, we give a brief summary of the SSA algorithm in 2D image decomposition, emphasizing the role of filters.
In Sec. 3, we give an example of the SSA decomposition of the image data, and the  symmetry of the eigenvector leads to the symmetry properties of the 
filters and these are reflected as the filter characteristics for the image data. 
An implication of these properties to the image denoising is briefly discussed. 
In Sec. 4, we summarize the results of this paper.
      
\section{ SSA Decomposition of 2D Image Data}  
\subsection{Basic algorithm}
Let us consider the 2D image data given by the matrix $A=\{ a_{ij} \}$ with $0 \leq i \leq M-1$ and $0 \leq j \leq N-1$. We adopt the moving window of any shape. For simplicity we assume here the $m \times n$ rectangular form $(m < M, n < N)$, which is shown as the box $[ \cdot \cdot \cdot ]$ in the following equation
\begin{eqnarray}
A= 
\left( 
\begin{array}{cc}
\left[ 
 \begin{array}{ccc}
  a_{00} & .. & a_{0,n-1}  \\
  .. & .. & .. \\
  a_{m-1,0} & .. & a_{m-1,n-1}  \\
  \end{array}
 \right]
  & 
  \begin{array}{ccc}
  a_{0,n} & .. & a_{0,N-1} \\
  .. & .. & .. \\
  a_{m-1,n} & .. & a_{m-1,N-1} \\
  \end{array} 
\\
  \begin{array}{ccc}
  a_{m,0} & .. & a_{m,n-1} \\
  .. & .. & .. \\
  a_{M-1,0} & .. & a_{M-1,n-1} \\
  \end{array}
& 
 \begin{array}{ccc}
  a_{m,n} & .. & a_{m,N-1} \\
  .. & .. & .. \\
  a_{M-1,n} & .. & a_{M-1,N-1} \\
  \end{array}
\\
\end{array} 
\right)  \ .
\end{eqnarray}
Then, by moving it $[ \cdot \cdot \cdot ]$ from left to right and from top to bottom, for example, the trajectory matrix $X$ can be defined as 
\begin{eqnarray}
X= 
\left( 
\begin{array}{cccccccc}
a_{00} & a_{10} & .. & a_{m-1,0} &.. &a_{0,n-1} & .. & a_{m-1,n-1} \\
a_{01} & a_{11} & .. & a_{m-1,1} &.. &a_{0,n} & .. & a_{m-1,n} \\
.. & .. & .. & .. & .. & .. & .. & .. \\
.. & .. & .. & .. & .. & .. & .. & .. \\
\end{array}
\right)   \  .
\end{eqnarray}
To introduce the filter for the 2D data, we consider an arbitrary $K=mn$ dimensional orthonormal vector $\{ v_j : j=1,...,K=mn \}$
. We arbitrarily assign the components of $v_j$ in the moving window, 
\begin{eqnarray}
 \left[ 
 \begin{array}{cccc}
  v_1 & v_{m+1} &  .. & v_{K-m+1}  \\
  .. & .. & .. & .. \\
  v_m & v_{2m} & .. & v_K  \\
  \end{array}
 \right] \ \ . \label{MovingWindow}
\end{eqnarray}
In the SSA, we adopt the vector set $\{ v^{(k)} : k=1,...,K=mn \}$ which is the eigenvectors of the matrix $X^TX$ and the filters are defined as, 
\begin{eqnarray}
 F^{(k)} = \left[ 
 \begin{array}{cccc}
  v^{(k)}_1 & v^{(k)}_{m+1} &  .. & v^{(k)}_{K-m+1}  \\
  .. & .. & .. & .. \\
  v^{(k)}_m & v^{(k)}_{2m} & .. & v^{(k)}_K  \\
  \end{array}
 \right] \ \ . \label{FilterF}
\end{eqnarray}
The following convolution-type linear relation express the forward filtering 
\begin{eqnarray}
a_{ij} & \longrightarrow & b^{(k)}_{ij} \nonumber \\ 
& = & v^{(k)}_1a_{ij}+v^{(k)}_2a_{i+1,j}+...+v^{(k)}_{K-m+1}a_{i,j+n-1} \nonumber \\
& & +...+v^{(k)}_{K}a_{i+m-1,j+n-1} \ \ .
\end{eqnarray}
Next, we perform the reverse filtering with the {\it point symmetric} filter 
\begin{eqnarray}
\left[ 
 \begin{array}{cccc}
  v^{(k)}_K & .. & v^{(k)}_{2m} & v^{(k)}_m  \\
  .. & .. & .. & .. \\
  v^{(k)}_{K-m+1} & .. & v^{(k)}_{m+1} & v^{(k)}_1  \\
  \end{array}
 \right] \ \ ,
\end{eqnarray}
for the reference point $a_{ij}$ with the relation
\begin{eqnarray}
b^{(k)}_{ij} & \longrightarrow & d^{(k)}_{ij} \nonumber \\ 
& = & v^{(k)}_1b^{(k)}_{ij}+v^{(k)}_2b^{(k)}_{i-1,j}+...+v^{(k)}_{K-m+1}b^{(k)}_{i,j-n+1} \nonumber \\
& & +...+v^{(k)}_{K}b^{(k)}_{i-m+1,j-n+1} \ \ .
\end{eqnarray}
We assume the periodic boundary for the image data $A$. 
In the SSA, the vectors $\{ v^{(k)} \}$ are determined as the eigenvector of the lag-covariance matrix $C \equiv X^TX/(L-K+1)$. It is easy to show that the completeness of the orthonormal eigenvectors leads to the perfect-reproduction property of the two-step filtering irrespective of the window shape,
\begin{equation}
 a_{ij}= {1 \over K} \sum_{k=1}^K d^{(k)}_{ij}  \ \ .
\end{equation}
The properties of the filters are clearly seen in the Fourier space. By embedding the eigenvector in the $M \times N$ matrix 
\begin{eqnarray}
V^{(k)}_{ij} =
\left( 
\begin{array}{cc}
\left[ 
 \begin{array}{ccc}
  v^{(k)}_1 & .. & v^{(k)}_{K-m+1}  \\
  & .. & \\
  v^{(k)}_m & .. & v^{(k)}_K  \\
  \end{array}
 \right]
  & 
  \begin{array}{ccc}
  0 & .. & 0 \\
  & .. & \\
  0 & .. & 0 \\
  \end{array} 
\\
  \begin{array}{ccc}
 0 & .. & 0 \\
  & .. & \\
 0 & .. & 0 \\
  \end{array}
& 
 \begin{array}{ccc}
  0 & .. & 0 \\
  & .. & \\
  0 & .. & 0 \\
  \end{array}
\\
\end{array} 
\right)   \ ,
\end{eqnarray}
we define the discrete Fourier transform as
\begin{equation}
\hat{A}_{\alpha \beta} =\sum_{m=0}^{M-1} \sum_{n=0}^{N-1} {\rm exp}[2\pi i (\alpha m/M+ \beta n/N)] a_{mn} \ ,
\end{equation}
and
\begin{equation}
\hat{V}^{(k)}_{\alpha \beta} =\sum_{m=0}^{M-1} \sum_{n=0}^{N-1} {\rm exp}[2\pi i (\alpha m/M + \beta n/N)] V^{(k)}_{mn} \ .
\end{equation}
Then they satisfy the normalization
\begin{equation}
{1 \over MN} \sum_{\alpha,\beta} |\hat{V}^{(k)}_{\alpha \beta}|^2 =1 \ , \label{FilterNorm}
\end{equation}
and the completeness relations (perfect reproduction in the language of digital filtering)
\begin{equation}
{1 \over K}\sum_{k=1}^K |\hat{V}^{(k)}_{\alpha \beta}|^2=1 \ . \label{FilterComplete}
\end{equation}
The above properties Eqs. (\ref{FilterNorm}) and (\ref{FilterComplete}) are satisfied for arbitrary orthonormal vectors $\{ v^{(k)} \}$. The SSA algorithm employs the eigenvectors of the eigenvalue equation 
for $X^TX$
\begin{equation}
X^TX v^{(k)} = \lambda_k v^{(k)} \ ,
\end{equation}
and this criterion means that the quantities 
\begin{equation}
v^{{(k)}T} X^TX v^{(k)} =\lambda_k  \ ,
\end{equation}
are maximized preserving the orthonormality of the vector $v^{(k)}$ [9]. 
In this sense, the SSA algorithm is the principal component analysis in the space of the lagged vectors. In the next subsection, we discuss the symmetry properties of the eigenvectors $\{ v^{(k)} \}$ and it's implication to the 2D image decomposition.

\subsection{Symmetry properties of the filters}
 We assign the vector components for the rectangular-shaped moving window with $ K = m \times n$ in a natural order as in Eq. (\ref{MovingWindow}).  We assume the periodic boundary condition for the image data. Then the lag-covariance matrix $C \equiv X^TX/(L-K+1) $ becomes to be the bisymmetric matrix, satisfying the relation
\begin{equation}
C_{i,j}=C_{K-i+1,K-j+1} \ \ .
\end{equation}
This can be seen as follows. The moving window in Eq.(\ref{MovingWindow}) slides from left to right and up to bottom sweeping the whole image data $A$.  The component $C_{i,j}$ of the matrix $C$ is calculated by picking and summing up the matrix elements of $A$, corresponding to the position $v_i$ and $v_j$ in the window, and this is the same for $C_{K-i+1,K-j+1}$ with $v_{K-i+1}$ and $v_{K-j+1}$.    
The location of the components $v_i$ and $v_{K-i+1}$ is just point symmetric about the center of the window. This also is the same for the relation between $v_j$ and $v_{K-j+1}$. Thus the relative location between $v_i$ and $v_j$ is the same as that 
between $v_{K-i+1}$ and $v_{K-j+1}$.  Because of this, the matrix element $C_{i,j}$ is equal to $C_{K-i+1,K-j+1}$.  This property holds for higher dimensional cases if we adopt the periodic boundary condition. For one dimensional case, a single-channel time series for instance, the matrix $C$ is a symmetric Toeplitz matrix.
 
Thus the matrix $C$ commutes with the exchange matrix $J$ 
\begin{equation}
[C,J]=0 \ ,
\end{equation}
and if there is no degeneracy for the eigenvalues of $C$, the eigenvectors $ \{ v^{(k)} \}$ are either symmetric or antisymmetric
\[
Jv^{(k)} = \pm v^{(k)} \ .
\]
This means that the filters generated with the eigenvectors $\{ v^{(k)} \}$ have corresponding symmetry. In particular, for the case of square form $m=n$ the matrices expressing the filter $F^{(k)}$
\begin{eqnarray}
F^{(k)} = \left[ 
 \begin{array}{cccc}
  v^{(k)}_1 & v^{(k)}_{m+1} &  .. & v^{(k)}_{K-m}  \\
  .. & .. & .. & .. \\
  v^{(k)}_m & v^{(k)}_{2m} & .. & v^{(k)}_K  \\
  \end{array}
 \right] \ , \label{FilterSquare}
\end{eqnarray}
are the centrosymmetric or skew-centrosymmetric matrices corresponding to the  symmetric $Jv^{(k)}=v^{(k)}$ or antisymmetric $Jv^{(k)}=-v^{(k)}$ eigenvectors, respectively.    

As an example, we have carried out the SSA decomposition of the image data 'Building' with rather small filter of square form $3 \times 3$.   The nine eigenvalues of the matrix $X^TX$ are ordered from larger to smaller ones and are shown in Fig. 1.
The five eigenvectors corresponding to the larger eigenvalues are shown in Table 1.  Here, the components of the eigenvectors are ordered in the $3 \times3$ filter $F^{(k)}$ as in Eq.(\ref{FilterSquare}). As seen, the eigenvectors $v^{(1)}$, $v^{(4)}$ and $v^{(5)}$ are symmetric, while $v^{(2)}$ , $v^{(3)}$ are antisymmetric. 
\begin{figure}[th]
\includegraphics[width=8cm,clip]{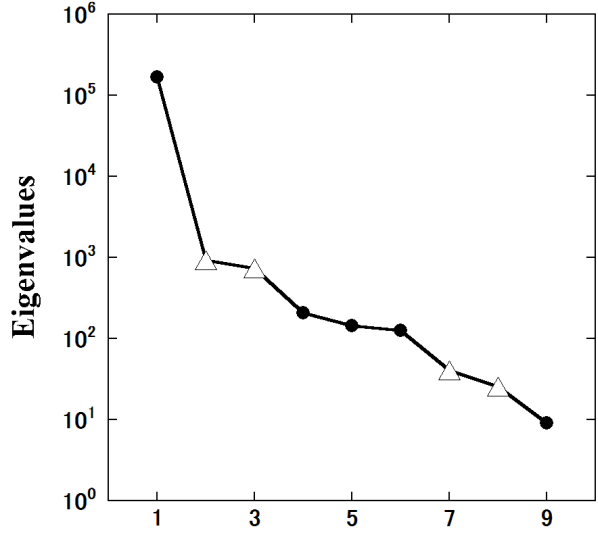}
\centering
\caption{The nine eigenvalues for the SSA decomposition of the image 'Building' with 
$3 \times 3$ filters.
The black circles(white triangles) are the eigenvalues corresponding to the  symmetric (antisymmetric) eigenvectors.}
\label{fig1}
\end{figure}

\begin{table}[th]
\caption{ The larger five eigenvectors of the matrix $X^TX$ for the SSA decomposition of the photo image 'Building' with $3 \times 3$ moving window. \label{tab1}}
\begin{tabular}{ccccc} \hline
 $v^{(1)}$ & $v^{(2)}$ & $v^{(3)}$ & $v^{(4)}$ & $v^{(5)}$  \\
\hline 
  0.3327 &  0.3747 & -0.4151 & -0.2671 &  0.0899 \\ 
 0.3338 &   -0.0197 & -0.4336 & 0.4223 & 0.2913 \\
 0.3326 & -0.4168  &  -0.3732 &  -0.2390 & 0.3057 \\ 
 0.3336 &  0.4307 & -0.0198 & -0.1967 & -0.4716 \\
 0.3348 &  0.0000 &  -0.0000 &  0.5558 & -0.4272 \\
 0.3336 & -0.4307 &  0.0198 &  -0.1967 & -0.4716 \\
 0.3326 &  0.4168 &  0.3732 &  -0.2390 &  0.3057 \\
 0.3338 &  0.0197 &  0.4336 &  0.4223 &  0.2913 \\
 0.3327 & -0.3747 & 0.4151 &  -0.2671 &  0.0899 \\
\hline
\end{tabular} 
\end{table}

To see the details of the decomposition characteristics, we have analyzed the filters as the differential operation. Let us assume that the image pixels are located on the lattice points $(ih,jk), \ i , j \in \bf{Z} $  with mesh length $h$ and $k$ for both $x$ and $y$ directions, respectively.  Here, the $x$-axis is taken for the horizontal direction from left to right and the $y$-axis for the vertical direction from up to bottom. The pixel value can be expressed with the function $f(x=ih,y=jk)$ on the lattice points, and the operation of the $3 \times 3$ filter $F^{(k)}$ can be expressed as

\begin{eqnarray}
f(x,y) & \longrightarrow & v^{(k)}_1f(x-h,y+k)+ v^{(k)}_2f(x-h,y) \nonumber \\ 
& + & v^{(k)}_3f(x-h,y-k)+ v^{(k)}_4f(x,y+k) \nonumber \\
 & + & v^{(k)}_5f(x,y)+ v^{(k)}_6f(x,y-k) \nonumber \\
& + & v^{(k)}_7f(x+h,y+k)+ v^{(k)}_8f(x+h,y) \nonumber \\
& + & v^{(k)}_9f(x+h,y-k) \ . \label{fx}
\end{eqnarray}
By carrying out the Taylor expansion we obtain 
\begin{eqnarray}  
f(x,y) & \longrightarrow & f(x,y)(\sum_{i=1}^9 v_i) \nonumber \\ 
& + & hf_x(-v_1-v_2-v_3+v_7+v_8+v_9) \nonumber \\
 & + & kf_y(v_1-v_3+v_4-v_6+v_7-v_9) \nonumber \\
 & + & h^2f_{xx}(v_1+v_2+v_3+v_7+v_8+v_9)/2 \nonumber \\
 & + & hkf_{xy}(-v_1+v_3+v_7-v_9)  \nonumber \\
 & + & k^2f_{yy}(v_1+v_3+v_4+v_6+v_7+v_9)/2 \nonumber \\
 & + & {\rm higher \  order  \  terms} \ , \label{fx_expand} 
\end{eqnarray}
where the vector components $v^{(k)}_j$ are abbreviated as $v_j$ and the $f_x \equiv  {\partial f(x,y) / \partial x}$ {\it etc.} 
The coefficients for each terms are shown in Table 2 for dominant five filters. The filters $F^{(k)}$ generated from the symmetric (antisymmetric) eigenvectors can be seen as the derivative filters of the even (odd) order which is simply a consequence of the centrosymmetric (skew-centrosymmetric) nature of the filters .

\begin{table}[th]
\caption{The coefficients appeared in Eq.(\ref{fx_expand})  for larger five eigenvectors.
 These clearly show how the filters work for the original image. \label{tab2}}
\begin{tabular}{ccccccc}\hline
 Filter & $f$ & $f_x$ & $f_y$ & $f_{xx}$ & $f_{xy}$ & $f_{yy}$  \\
\hline
     1  &   3.000 &    0.000 &  0.000 &  0.999 &  -0.000 &  0.999 \\
     2  &  0.000 & 0.123 &  2.444 & 0.000 & 0.000 & 0.000 \\
     3  &  0.000 &  2.444 &  -0.123 &  0.000 &  0.000 &  0.000 \\
     4  & -0.005 & 0.000 &  0.000 & -0.084 &  0.056 & -0.703 \\
     5  &  0.004 & 0.000 & 0.000 &  0.687 & 0.432  & -0.076　\\
\hline
\end{tabular} 
\end{table}

\begin{figure}[th]
\centering
\includegraphics[width=14cm,clip]{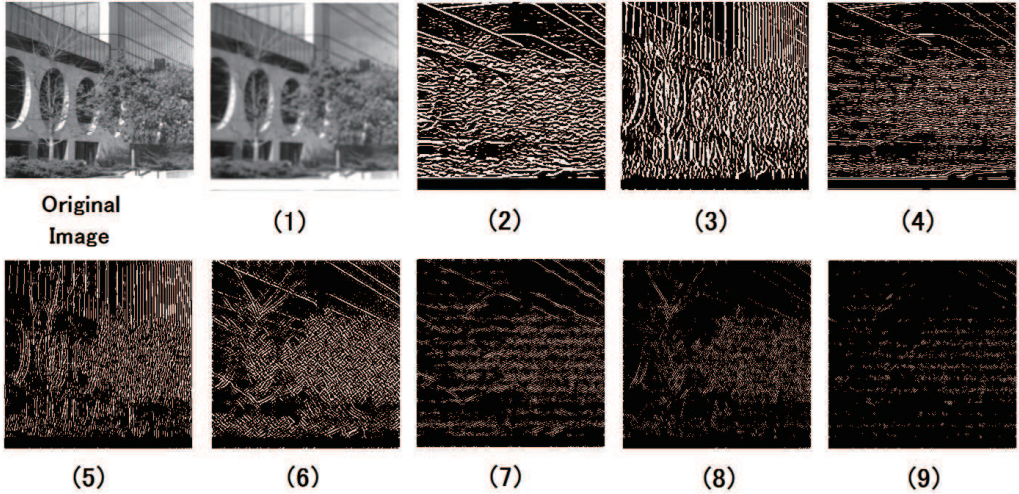}
\caption{The original image (upper left) and its SSA decomposition through $3 \times 3$ filters for the photo image 'Building'. The images $(k)$ are obtained by the two-step filtering with $F^{(k)}$.}
\label{fig2}
\end{figure}
The dominant filter $F^{(1)}$ works as smoother of the pixel values. The second and the third filters $F^{(2)}$ and $F^{(3)}$ are generated from the antisymmetric eigenvectors and work as the differential filter in $y$ and $x$ directions, respectively.  The filters $F^{(4)}$ and $F^{(5)}$ are generated from the symmetric eigenvectors and the dominant components are the  second derivative for various directions. 

The image decomposition for the 'Building' are shown in Fig. 2. We can see that the results of the decomposition reflect the characteristics of each filter.  For the decompositions  (2) and (4), the horizontal lines are enhanced corresponding to the large $f_y$,$f_{yy}$ components as in Table 2.  On the other hand, the vertical components are enhanced in the image (3) and (5).  As seen, each filter has it characteristics and the SSA decompositions reflect their characteristic properties.   The images $(7) \sim (9)$ correspond to smaller eigenvalues and they exhibit somewhat featureless pattern due to the large high-frequency components. These nine images are added to reproduce the original one in the SSA algorithm.

\subsection{SSA decomposition of images with noise}
To see how the SSA algorithm works for the decomposition of the image data with noise, we adopt an example 'Lenna' with $256 \times 256$ pixels. Each pixel takes the values $0 \sim 255$. The Gaussian noise $13.3N(0,1)$ is added for every pixels.  We have carried out SSA decomposition with a square filter of the size $11 \times 11$. 
The larger fifteen eigenvalues of the matrix $X^TX$ are shown in Fig. 3, where the black circles (white triangles) represent the eigenvalues corresponding to the symmetric (antisymmetric) eigenvectors.

\begin{figure}[th]
\centering
\includegraphics[width=7cm,clip]{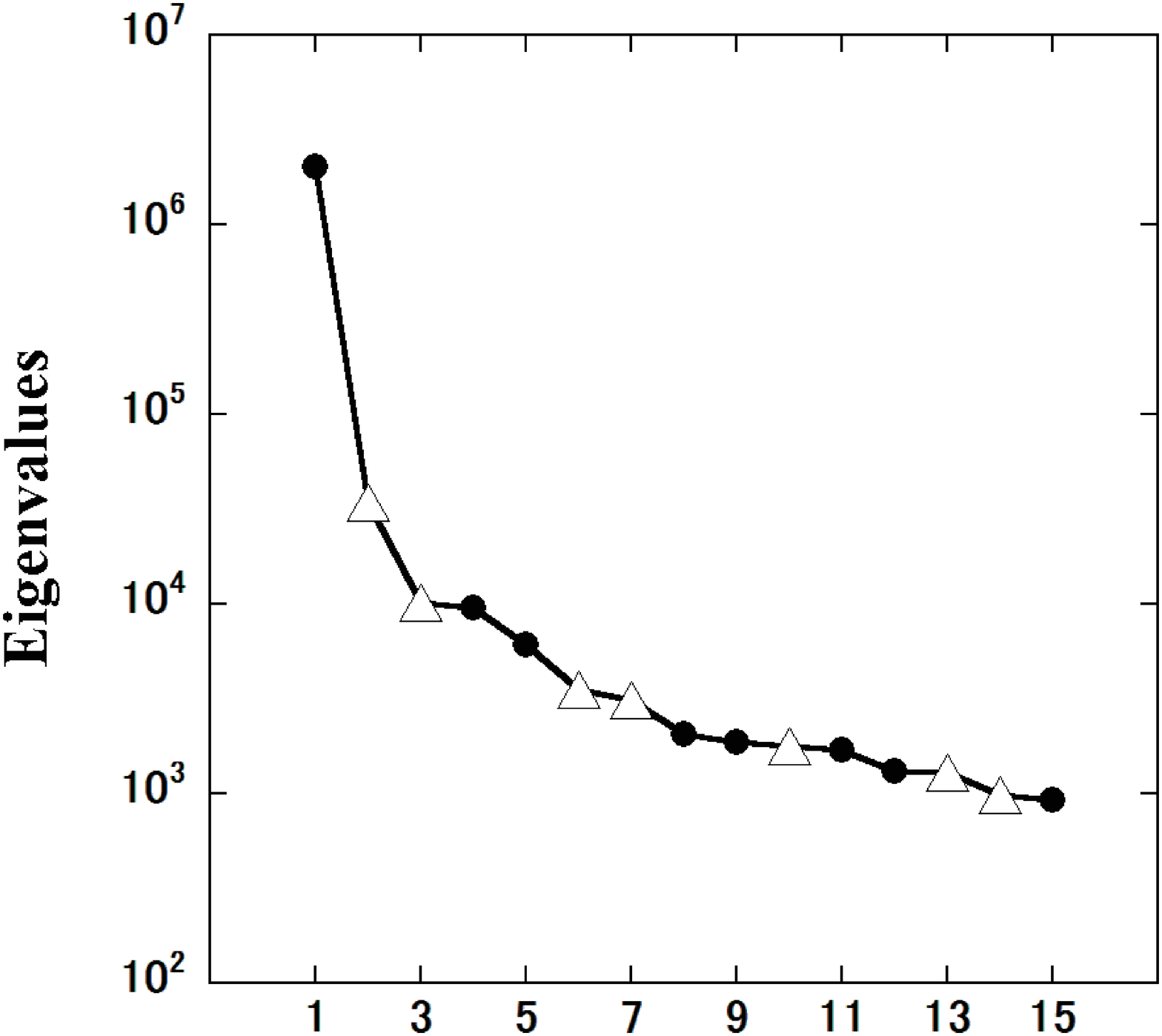}
\caption{The larger fifteen eigenvalues of $X^TX$ for the SSA decomposition of the image 'Lenna' with noise. The black circles (white triangles) are the eigenvalues with symmetric(antisymmetric) eigenvectors.}
\label{fig3}
\end{figure}

Similar to the case of the 'Building' image, we have examined how each filter works for the image by examining the coefficients in Eq. (\ref{fx_expand}) extended for the case of $11 \times 11$ filter. In this case, since the filter size is large, we have replaced the mesh lengths $h$ and $k$ by $h/3$ and $k/3$, respectively as,
\begin{equation}
f(x,y) \longrightarrow v_1^{(k)} f(x-5h/3,y+5k/3)+v_2^{(k)} f(x-5h/3,y+4k/3)+....  \ \ .
\label{expand}
\end{equation}
The coefficients $f$, $hf_x$, $kf_y$ {\it etc.} are shown in Table 3. 
\begin{table}[th]
\caption{The coefficients appeared in Eq.(\ref{fx_expand})  for larger five eigenvectors for the 'Lenna' with noise. In this case, since the filter size is large, we have replaced the mesh lengths $h$ and $k$ by $h/3$ and $k/3$ as in Eq.(\ref{expand}). The filters generated from the symmetric (antisymmetric) eigenvectors work as the derivative filters of the even (odd) order.
\label{tab3} }
\begin{tabular}{ccccccc} \hline
 Filter & $f$ & $f_x$ & $f_y$ & $f_{xx}$ & $f_{xy}$ & $f_{yy}$  \\
\hline
     1  &   11.000 &    0.000 & 0.000 &  6.079 &  -0.021 &  6.101 \\
     2  &  0.000 & 11.337 &  1.672 & 0.000 & 0.000 & 0.000 \\
     3  &  0.000 &  -1.669 & 11.133 &  0.000 &  0.000 &  0.000 \\
     4  & 0.067 & 0.000 &  0.000 & 5.092 &  2.654 & 0.555 \\
     5  &  0.005 & 0.000 & 0.000 &  -1.245 & 11.140  & 0.667　\\
\hline
\end{tabular} 
\end{table}

For the 'Lenna' image with noise, the filter characteristics are quite similar to the case of  the 'Building' photo image. The first filter $F^{(1)}$ works as the smoother for the original image. The smoothing operation works stronger than the case of 'Building', since the filter size is larger $11 \times 11$ in this case. The second and the third filters work as the differentiation to the $x$ and $y$ directions, respectively.  For the fourth and the fifth filters $F^{(4)}$ and $F^{(5)}$, the expansion coefficients $f_{xx}$ and $f_{xy}$ are large as seen in Table 3.    
 
\begin{figure}[th]
\center
\includegraphics[width=14cm,clip]{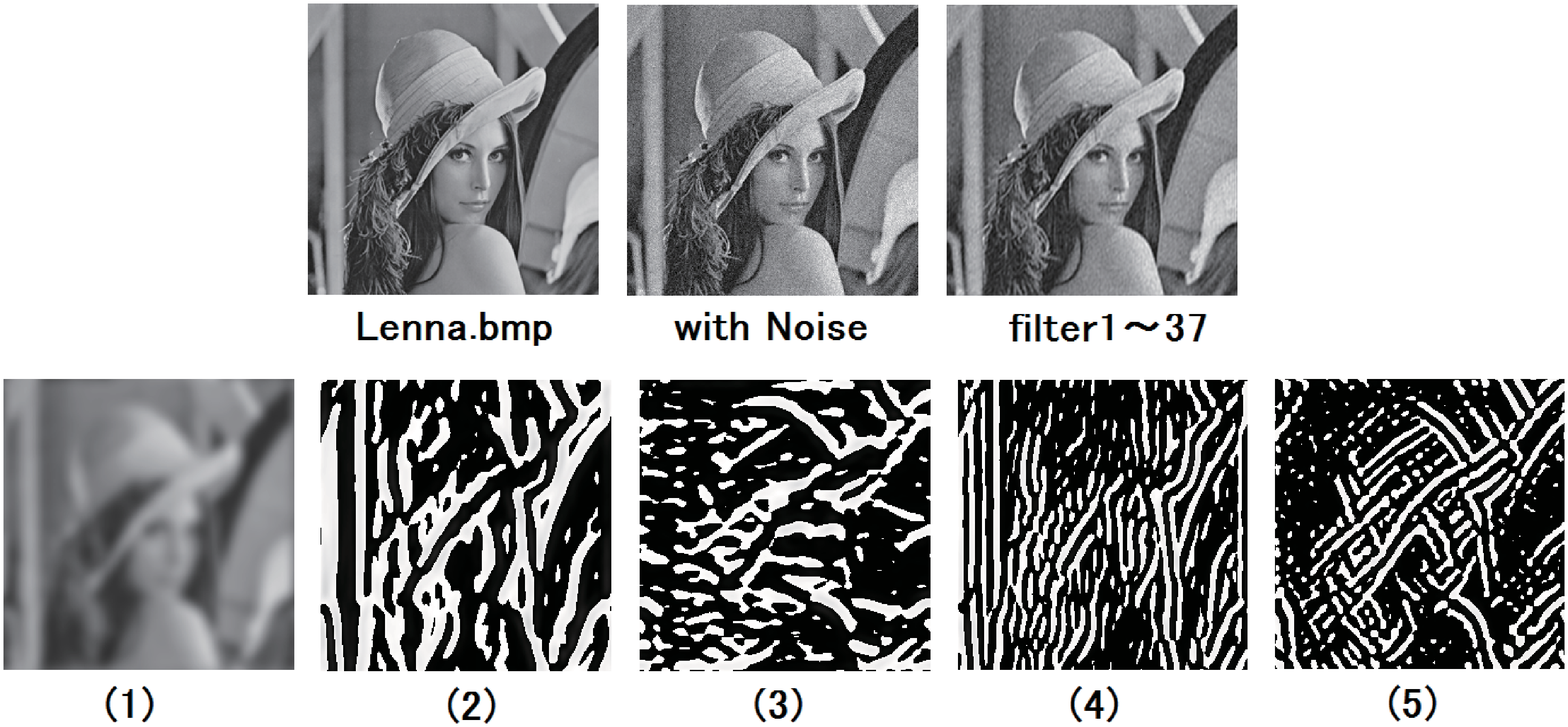}
\caption{Upper left and central images are the 'Lenna' with and without Gaussian noise.  The image is decomposed into $11 \times 11$ components with SSA algorithm. The upper right image is the partial sum of the SSA decomposition $k=1 \sim 37$. The lower images are the dominant five decompositions.}
\label{fig4}
\end{figure}

For the filters corresponding to the smaller eigenvalues, the derivative components are fragmented and the filtered images gradually become to be structureless including the    the higher frequency components.

In Fig. 4, we have shown the 'Lenna' image with and without noise together with the dominant five SSA decomposition images with noise. Corresponding to the filter characteristics as seen in the Table 3, we can see that the edge lines in various directions are enhanced and are extracted from the original image. For the images (2) and (3), the vertical and the horizontal lines are enhanced. 
The filters $F^{(4)}$ and $F^{(5)}$ have large $f_{xx}$ and $f_{xy}$ components, and hence the finer details of the edges in the vertical or the diagonal lines are enhanced for the images (4) and (5).

\begin{figure}[th]
\center
\includegraphics[width=8cm,clip]{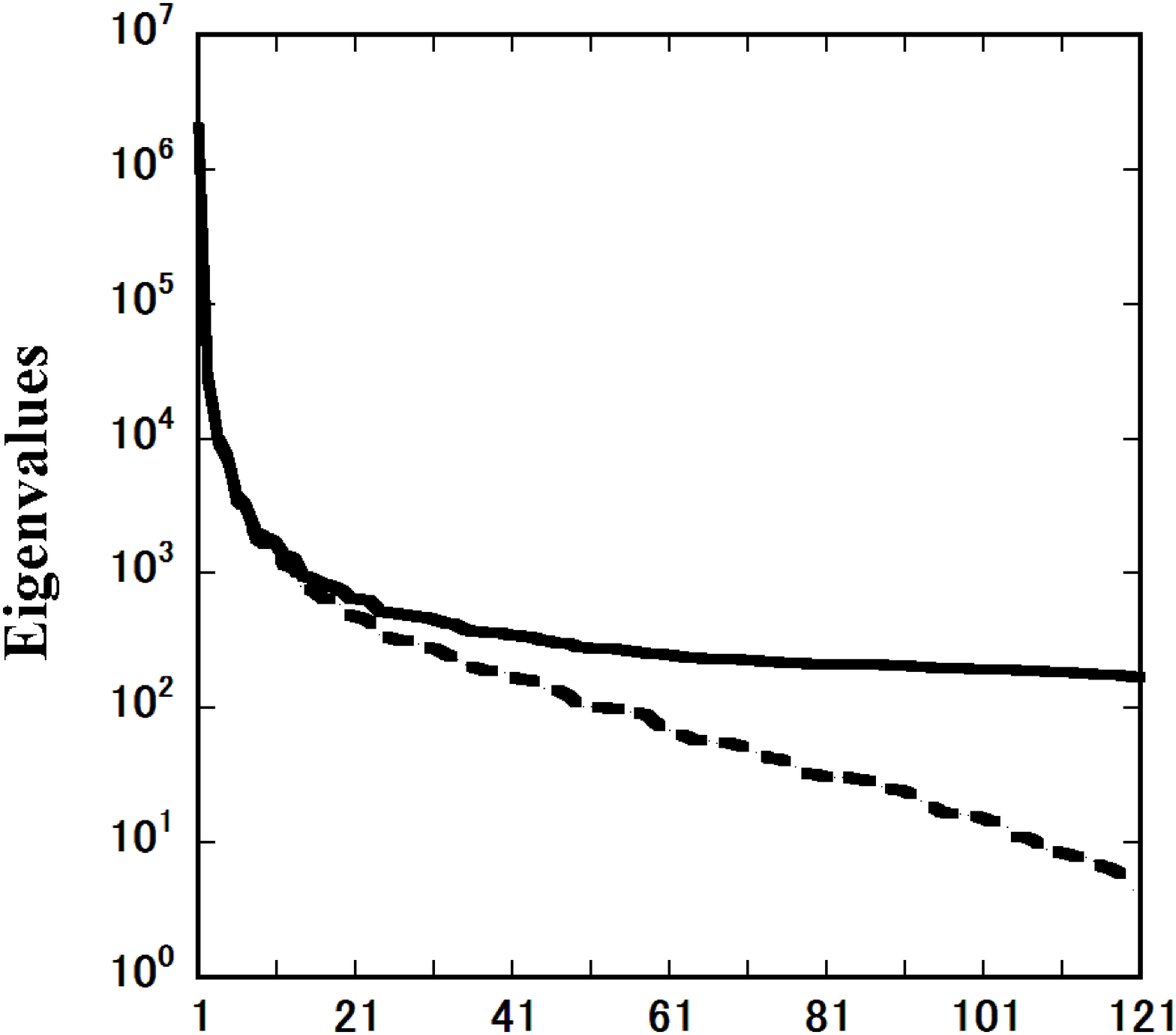}
\caption{The $11 \times 11$ eigenvalues are shown for the SSA decomposition of the image 'Lenna'. The dashed and the solid lines are the results with and without the Gaussian noise.}
\label{fig5}
\end{figure}

In order to see the relation between the SSA decomposition and the noise components, we show all the 121 eigenvalues for the 'Lenna' image with and without the Gaussian noise in Fig.\ref{fig5}. The added noise enhances the smaller eigenvalues coming from the 
high-frequency components. 

\begin{figure}[th]
\centering
\includegraphics[width=7cm,clip]{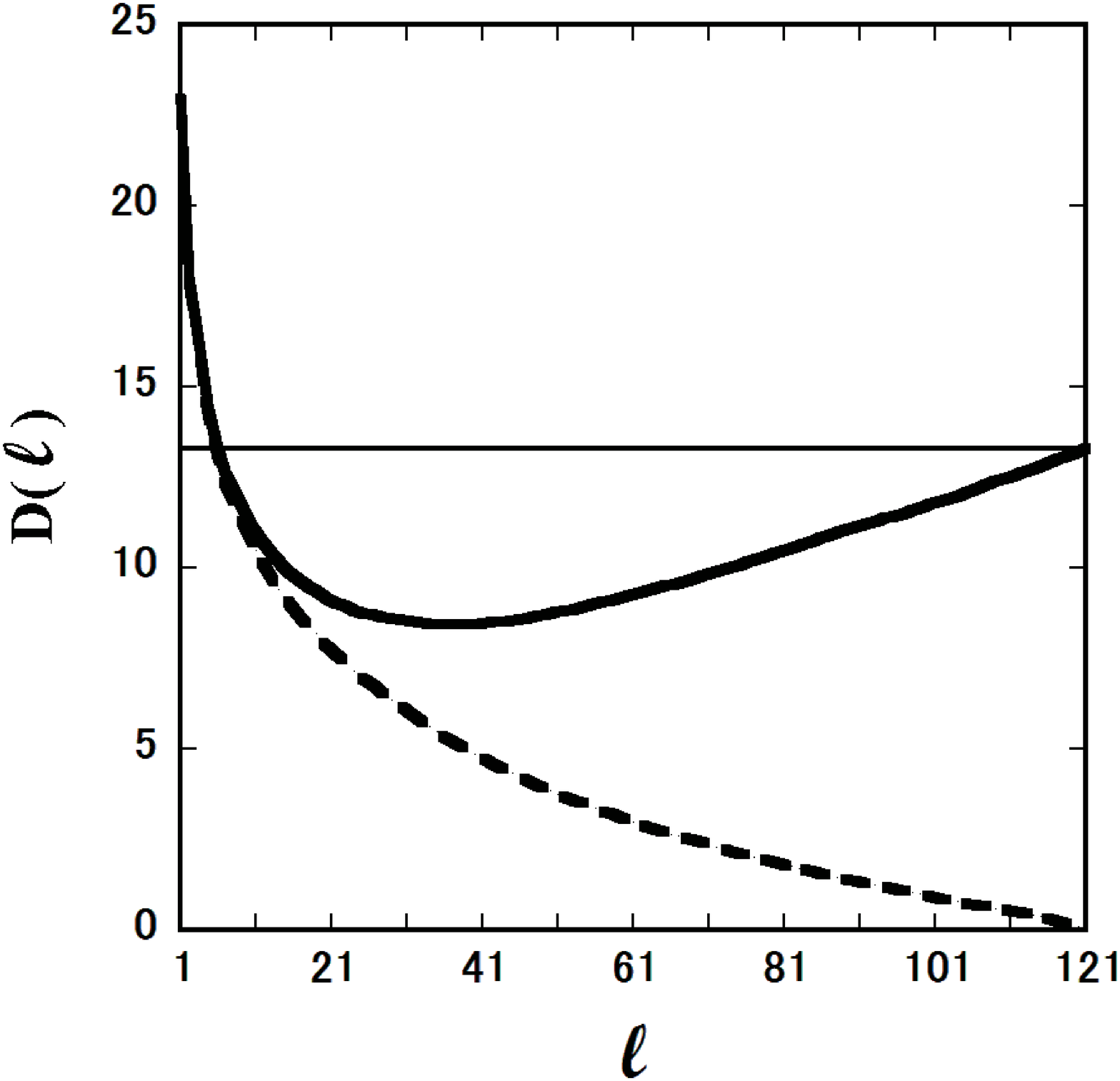}
\caption{The RMS distance $D(\ell)$ between the original image and the partial sum of the
 SSA decomposition $k=1 \sim \ell$ as in Eq.(\ref{RMSdistance}). The solid and the dashed lines are the results with and without Gaussian noise $13.6N(0,1)$.  The horizontal line represents the RMS distance between the noiseless and noisy images.}
\label{fig6}
\end{figure}
We have also calculated the RMS distance $D(\ell)$ 
between the original noiseless image $a_{ij}$ and the partial sum of the SSA
decomposition $d^{(k)}_{ij}$ \cite{Rodriguez-Aragon}, 
\begin{equation}
 D(\ell)=\sqrt{ {1 \over MN} \sum_{i,j} ( a_{ij} - {1 \over K}\sum_{k=1}^{\ell} d^{(k)}_{ij} )^2 } \ . 
\label{RMSdistance}
\end{equation}
The results are shown in Fig. \ref{fig6}. Obviously, for the noiseless case, the distance converges to zero as $\ell \rightarrow 121$, which merely means that the sum of the SSA decomposition reproduce the original image. For the case with noise term, the RMS distance decreases as $\ell : 0 \rightarrow 37$, while it increases as $\ell :37 \rightarrow 121$. The RMS distance takes minimum value at $\ell = 37$. We could discard the components  $\{ d^{(k)}; k > 37 \}$ in order to obtain better images. The image corresponding to the partial sum $k=1 \sim 37$ is also shown in Fig. \ref{fig4}. The RMS distance between noiseless and noisy images is $D(\ell=121)=13.3$ which is just the standard deviation of the added noise. The RMS distances for the partial sum of the SSA decomposition ${k=1 \sim 37}$ 
 with and without noise are $D(\ell=37)|_{\rm{with \ noise}}=8.42$ and $D(\ell=37)|_{\rm{noiseless}}=5.21$, respectively. 

So far, much efforts have been devoted to the image restoration. Various linear or non-linear filtering operations have been examined to improve the noisy images [1,5]. The performance of the filtering operation depends on the properties of the image itself or the nature of the noise. 
In the SSA, the noisy images are exactly decomposed into arbitrary number of components and each image decomposition has its own characteristics. Thus, by combining the conventional nonlinear denoising algorithms and SSA decomposition, we expect to
develop better denoising algorithms. Since the main purpose of this paper is to show the characteristic feature of the SSA decomposition of the 2D image data, we do not pursue these points further.  

%

\section{Summary and Conclusions}
The filtering interpretation of the SSA algorithm enabled us to study the 
 characteristics of the image data decomposition in detail. The periodic boundary condition and the rectangular moving window leads to the bisymmetric lag-covariance matrix $C$ and its eigenvectors are either symmetric or antisymmetric.  For the symmetric (antisymmetric) eigenvectors, the corresponding filter $F^{(k)}$ are  centrosymmetric (skew-centrosymmetric) matrices and they are interpreted as the derivative filter of even (odd) order. The SSA algorithm can be understood as the optimal and adaptive generation of the filters $F^{(k)}$ and the two-step point-symmetric operation of these filters to the reference points of the original image.  The two-step filtering ensures the reality of the filters in the Fourier space and it introduces no distortion or shift to the data. Also the completeness of the eigenvectors ensures the perfect reproductivity of the SSA decomposition.
Each image components in the SSA decomposition exhibit 
characteristic features, which comes from the properties of the adaptive filters generated from the eigenvectors of the lag-covariance matrix. We briefly discussed the relation of the SSA decomposition and the noise reduction of the images. Much efforts have been devoted so far for the image denoising. We expect that the filtering approach for SSA algorithm enables us to develop an novel denoising algorithm by combining the SSA decomposition with conventional nonlinear filtering algorithms. The study in this direction is now under way.

\noindent
{\bf Acknowledgment} \\
\noindent
This work is supported by Grant-in-Aid for Scientific Research from the Japan Society for the Promotion of Science; Grant Numbers (24650150) and (26330237).

\end{document}